Analysis of Dutch Master Paintings with Convolutional Neural Networks

Steven J. Frank and Andrea M. Frank

In less than a decade, neural networks have achieved widespread deployment in applications ranging from self-driving cars to real-time language processing to medical diagnostics. The rate and volume level at which new potential uses are announced obscures how few human workers have actually been displaced. For applications whose value is high enough to justify the cost of research and development, the risks of taking human expertise out of the loop may be too great. For applications involving creativity and subjective judgment, the uncertainties may be irreducible.

Analysis of artwork represents one such creative domain. Neural networks have been used to categorize art images by movement and style (Lecoutre et al., 2017; Balakrishnan et al., 2017; Tan et al., 2016; Saleh et al., 2015; Bar et al., 2014), to classify paintings by artist rather than style (van Noord et al., 2015), and to extract and analyze artists' brushstrokes (Li et al., 2012). These techniques have achieved varying degrees of success but few efforts have addressed the more difficult question of attribution and forgery detection. The reason for this stems from some fundamental limitations of neural networks. Organized in a highly interconnected brain-like fashion, they can analyze and recognize patterns in a wide range of complex input — an image, spoken words, stock prices, or weather data, for example. But a neural network must be trained for a given task. Convolutional neural networks (CNNs) have computational architectures particularly suited to classifying images; all of the art-related research efforts mentioned above utilize CNNs. During training, a CNN processes many labeled images relevant to the classification (e.g., whether an image contains a cat), and gradually adjusts the internal components that analyze the pixels of each image. The ability to perform the task depends critically on the quality of training, and in this regard, the overwhelming criterion of quality is quantity: to approach human levels of discriminatory power, a CNN must be trained on thousands if not millions of images. Unfortunately, while the supply of images containing cats is (seemingly) inexhaustible, the same is not true of Vermeers, van Goghs, and Rembrandts.

A second problem is image size. Most CNNs process images smaller than 300 × 300 pixels. The complexity of processing scales with the pixel count of an image, so as the image size grows, computational and memory resources can quickly become overwhelmed and training can slow to a near halt. The conventional expedient of downsampling the image sacrifices the fine or subtle image features that distinguish authentic works from forgeries.

*Methodology and Scope: Analysis of Artwork Using Image Entropy*

Our approach, as we describe in Frank et al. (2020), is to divide images into tiled segments and use, for training and prediction, tiles that satisfy a visual information criterion. In particular, we utilize image "entropy," a concept from information theory that captures visual diversity. A low-entropy image, such as a simple bar or pattern, contains little unique visual information; there is much redundancy and the image may be easily compressed into a simplified format. A high-entropy image, such as a Dutch still life painting, contains substantial visual diversity and complexity. Whereas the absolute amount of visual information tends to scale with image size, the entropy of a tile is largely independent of size; even a small but visually busy tile can have high entropy. After we decompose an image into square tiles, we discard the ones whose entropies do not equal or exceed the overall entropy of the source image. Visual diversity, we find, drives CNN performance, and these stand-out tiles tend to stand in well for the source image yet are small enough to be processed efficiently. A single high-resolution image can yield hundreds of usable tiles. It is therefore possible, even with relatively few source images, to successfully train a CNN using thousands of tiles that individually retain fine image features.

A well-curated set of training images is critical to success. Predictive capability depends on the ability to distinguish an artist's work not only from the images in the training set, but from new images that may differ subtly in style or execution from anything previously presented to the CNN. The training set must therefore contain images of varying degrees of similarity to the artist whose work is to be studied, and some of those images must be closely similar.

We have used our system to study portraits by Rembrandt Harmenszoon van Rijn (1606-1669) and landscape paintings by Vincent Willem van Gogh (1853-1890). We chose these artists and genres for their obvious pictorial differences in order to test the versatility of our approach. As we will see, however, their further differences have more significant and interesting implications.

Most importantly for our purposes, van Gogh worked largely if not entirely alone while Rembrandt ran an active (and profitable) workshop populated by tuition-paying students seeking to learn from the master. Rembrandt's students copied his works and served as his assistants. Many, such as Govaert Flinck, Ferdinand Bol, and Carel Fabritius, became recognized artists in their own right. Their sheer number — 100 or so — and Rembrandt's working habits have complicated scholarship of his works for hundreds of years. One authority notes that "it seems most likely that Rembrandt, like Rubens in Antwerp and Van Dyck in England, used studio assistants to help him produce paintings for the market, especially during the 1630s when his work was in great demand" (Wheelock, 2014). Indeed, experts have long suspected that Rembrandt included drawings by his students along with his own in albums sold as his work (Finkel, 2009). Rembrandt also signed his students' work on occasion, enhancing the profit he made on its sale (Scallen, 2003).

In 1968 the Rembrandt Research Project (RRP) was formed to answer the many persistent attribution questions with some finality. They did so energetically, rejecting dozens of works. When the Dutch art historians of the RRP disagreed over stylistic criteria, which was frequently, the very existence of such disagreement often resulted in de-attribution. Rejected works included a signed 1637 self-portrait and the Frick Collection's beloved *Polish Rider*.[1] A century ago Rembrandt's total output was estimated at 711 works, but by 1989, only 250 works had survived the RRP's judgment. Although the committee later restored approximately 90 works to the canon before disbanding in 2011, many Rembrandt paintings remain mired in controversy. *The Polish Rider* (which the Frick dates ca. 1655) is now recognized as a Rembrandt, but another painting, *The Man with the Golden Helmet* (Gemäldegalerie, Berlin), is among those that have bounced from attribution to de-attribution, with the current scholarly consensus supporting its demotion to "circle of Rembrandt."

The RRP, in its deliberations, occasionally considered the possibility of student contributions in various works, but absent contemporaneous records or clear stylistic distinctions, concluded that "with only one or two exceptions … as a rule one and the same hand did produce the whole of the painting" (Bruyn et al., 1982). More recently, some scholars, noting the doubts cast by postmodern philosophy on traditional notions of authorship, question

---

[1] RRP member and leading Rembrandt scholar Ernst van de Wetering more recently characterized the RRP's initial verdict as "doubt" rather than de-attribution. Van de Wetering (2017).

whether the traditional presumption of unique authorship is a romantic, and frequently mistaken, notion:

> The conception of painting as autographic art, fully executed by the painter's own hand, and enabling an authentic work to be distinguished from a copy, is essential in defining the work of art, and certainly is a major premise in art history discourse. Today, this traditional view is being called into question by studies based on sources revealed by social history since the 1970s (and by the new, later data derived from scientific study of the paintings). These sources attest to broad types of collaboration among artists, from the Renaissance onward, and highlight the importance of the many and diverse variations produced by artists of their own work and that of others. (Guichard, 2010.)

Many contemporary artists outsource fabrication of their works to ateliers or specialized production facilities without adverse impact on marketplace value; collectors of such work appear satisfied with an authorship construct limited to the artist's vision and its supervised fulfillment. Between the sole authorship of van Gogh and the absent hand of Jeff Koons, however, lies a broad range of artist collaborations — visible and invisible, known and hidden — among contemporaries, with students, and in the form of directions given to anonymous assistants. The ability to identify the different authors of a work of art, or at least to distinguish between the hand of the master and the work of others, would provide significant insight into an artist's methods and a deeper understanding of his or her work.

It is to this task that we have applied our neural network system. Besides illuminating distinct authorships, our analysis can also reveal areas of restoration and later revision. As will hopefully become clear, however, automated analysis of as human an endeavor as artistic production is inevitably uncertain. Absent supporting scholarship from traditional sources, our approach is to limn the often intriguing possibilities rather than state a conclusion.

### *Neural Network Analysis of Rembrandt Portraits*

Trained on a curated dataset as described in Frank et al. (2020), our CNN distinguishes between Rembrandt's work and portraits of similar style painted by others (including his students) with perfect accuracy on a 24-image test set when classification probabilities are averaged over an entire painting. We tested six works whose attributions have changed over time and our CNN's classifications accord with the current scholarly consensus in all but one case — and as discussed below, we modestly question that current attribution.[2]

| Title | Scholarly Consensus | Our Classification | Probability |
|---|---|---|---|
| *Man with the Golden Helmet* | School of Rembrandt | Rembrandt | 0.85 |
| *Portrait of A Young Gentleman* | Rembrandt | Rembrandt | 0.81 |
| *Portrait of Elisabeth Bas* | Ferdinand Bol | Not Rembrandt | 0.39 |
| *The Polish Rider* | Rembrandt | Rembrandt | 0.57 |
| *Portrait of a Man ("The Auctioneer")* | Follower of Rembrandt | Not Rembrandt | 0.29 |
| *Portrait of A Young Woman (Kress Collection/Allentown Art Museum)* | Rembrandt | Rembrandt | 0.82 |

Peak accuracy for Rembrandt occurs at a tile size corresponding roughly to the size of the face in a typical portrait, meaning that each tile examined represents 3% to 4% of the overall area of the painting. This does not mean, however, that the CNN ignores smaller-scale features; rather, it processes features on all scales up to the tile boundaries. A face-sized tile allows the CNN to make classifications based on visual structures ranging from fine brushstrokes to larger compositional elements. Using smaller tiles that confine analysis *only* to brushstroke-level detail, by contrast, degrades performance to little better than guessing — for Rembrandt. The opposite is true for van Gogh, as we discuss below.

The fraction of tiles that survive the sieve of our entropy criterion is small, averaging less than 15% — too few to support a reliable classification. We therefore allow tiles to overlap both horizontally and vertically, which drastically increases their number and reduces classification error commensurately. Our classifications are not binary but rather take the form of probabilities between zero and one with a "decision boundary" at 0.5: values equal to or exceeding the decision boundary correspond to classification as Rembrandt, while values below the decision boundary represent a non-Rembrandt classification. We define classification error as the absolute difference between the average tile probability and 0.5 for a misclassified painting.

---

[2] The works and their associated probabilities were *Man with the Golden Helmet* (school of Rembrandt), 0.88; *Portrait of a Young Gentleman* (Rembrandt), 0.65; *Portrait of Elisabeth Bas* (Rembrandt's student Ferdinand Bol), 0.17; and *The Polish Rider* (Rembrandt), 0.92.

The relatively small size of our dataset was dictated, in part, by the need to select paintings that have been subjected to sufficient scholarly scrutiny to be attributable, to the utmost degree of certainty, to Rembrandt's sole hand — otherwise the basis of our training will be flawed as will our CNN's classifications. This is but one irreducible source of possible error that limits the certainty of any conclusions. It is for this reason that, when assigning probabilities to regions within a painting, we limit ourselves to four ranges; finer classifications might imply greater precision than can be justified. In particular, we explore the regions of a painting important for classification — i.e., the regions where our entropy criterion was satisfied — using "probability maps," which color-code the probabilities assigned to the examined regions of an image using four colors: red corresponds to high-likelihood ($\geq 0.65$) classification as Rembrandt; gold to moderate-likelihood ($0.5 \leq p < 0.65$) classification as Rembrandt; green to moderate-likelihood ($0.5 > p > 0.35$) classification as not Rembrandt; and blue to high-likelihood ($\leq 0.35$) classification as not Rembrandt. Gray regions of a probability map represent tiles that did not pass the image-entropy selection criterion and were not examined.

Rembrandt painted in different styles and genres, ranging from theatrical religious and historical subjects to quiet landscapes and formal portraits. We focused on portraits. The dataset we employed to train our CNN for Rembrandt included high-resolution images of portraits he painted from the early 1630s nearly until his death in 1669. The other 50% of our training set consisted of portraits by other artists, selected for varying degrees of pictorial similarity to the Rembrandts. The objective in identifying the comparative non-Rembrandt images was, once again, to train the neural network to make fine as well as coarse distinctions and to generalize beyond the training images. The paintings in the training set include works that had once been attributed to the master himself, but which are now qualified as "school of," "workshop of" or "circle of" Rembrandt; portraits attributed to students of Rembrandt (Govaert Flinck, Carel Fabritius, and others); and works by Dutch contemporaries of Rembrandt, including several (like Frans Hals) whose style is easily distinguishable from Rembrandt's, even to a non-specialist.

Our Rembrandt dataset was not large — about half the size of the one we used for our van Gogh study — which allowed us to steer clear of controversial attributions and also to assess whether, trained on a small number of paintings, our CNN could nonetheless produce reasonably accurate classifications. The extant work of many Old Master artists is quite limited, yet this does not deter forgers: only 36 authentic paintings by Johannes Vermeer have been identified, for example, but his work was famously forged by Han van Meegeren during the Nazi era. The performance we achieved on a limited training set demonstrates the versatility of our approach for studying very scarce artwork.

The intersection of computational analysis and traditional art scholarship, and the criticality of the latter to the former, is exemplified by the probability map generated for the *Portrait of Johannes Wtenbogaert* (1633; Rijksmusum, Amsterdam).

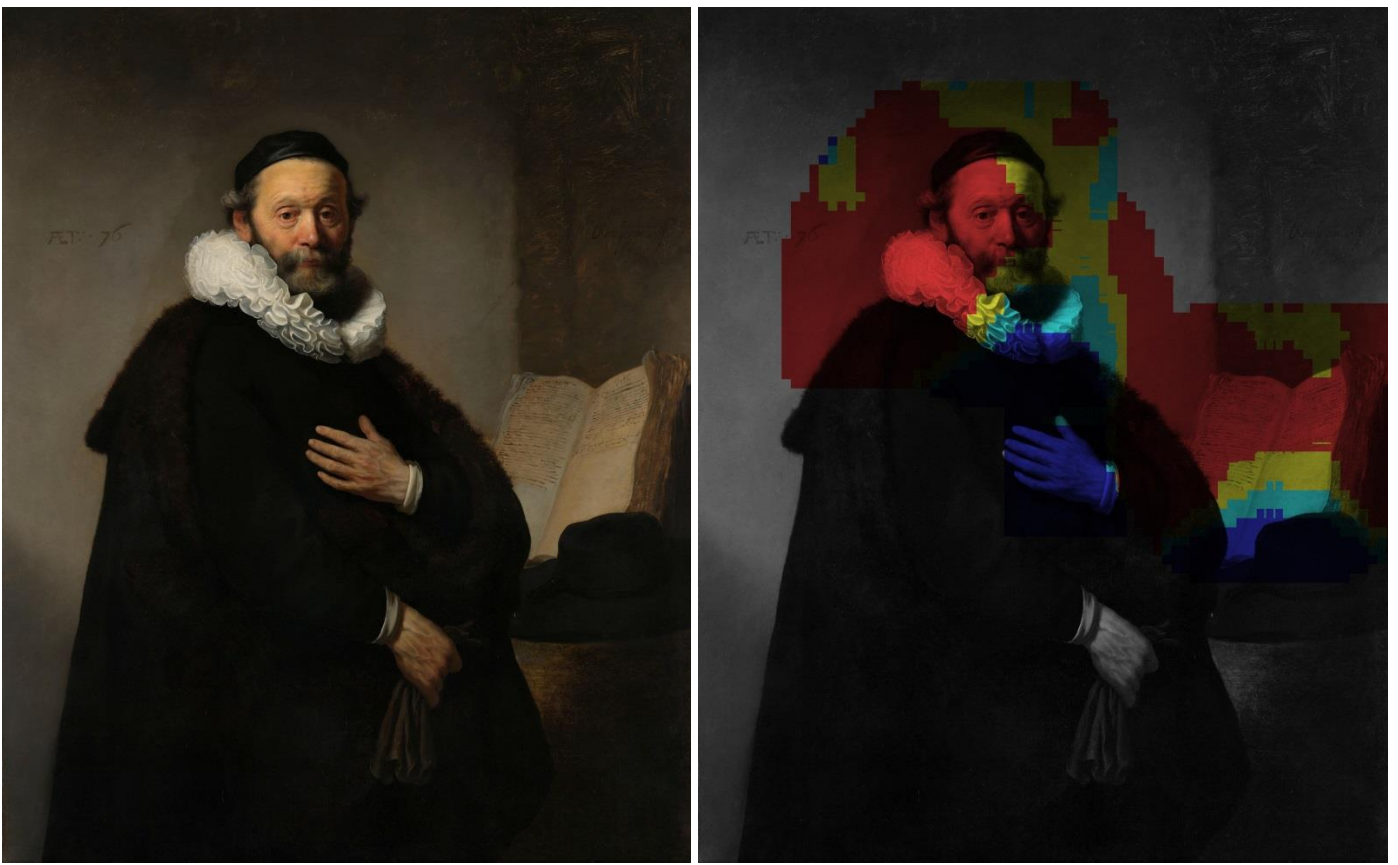

Fig. 1: *Portrait of Johannes Wtenbogaert, Probability map*

The 76-year-old preacher, a revered figure and champion of tolerance in the Netherlands, is depicted in a simple, scholarly setting with a dark, largely undetailed cloak that accentuates the subject's face and elaborate collar. Rembrandt's characteristic three-quarter lighting dramatically highlights the preacher's furrowed features and glistening eyes. It is noteworthy, initially, how relatively little of the painting our CNN actually examined; much of the cloak and background lack sufficient visual diversity to satisfy the entropy criterion. Nonetheless, the analysis covers enough of the canvas to support a classification, and our CNN reports an overall probability of 0.57 — that is, a weak Rembrandt classification. The regional distribution of probabilities shows that the face, part of the elaborate collar, and most of the open book are classified as Rembrandt; a portion of the collar and the lower area of the open book are classified as not by Rembrandt. Most prominently, the topmost hand is colored dark blue, corresponding to a strong likelihood that it did not originate from the master's brush.

In his most recent treatise, leading Rembrandt scholar Ernst van de Wetering attributes *Johannes Wtenbogaert* to "Rembrandt and workshop" (van de Wetering, 2017) and notes, "Given the apparent weaknesses in the execution of the hands, Rembrandt must have left the painting of these parts to an assistant in the sitter's absence." This accords with the dark blue probability assignment, which encompasses examined hand and bleeds into a portion of the collar. Van de Wetering notes that the portrait's subject sat for no longer than a single day, likely prompting the sitter to follow then-common practice and leave his intricate collar with Rembrandt for later rendering. Although it is tempting to ascribe the blue portion of the collar to a student's precocious efforts to tackle a pictorial challenge, more likely it represents a spurious artifact of the tile size. The filigreed pixelation of the probability map is formed by significantly overlapping tiles, each of which is fairly large for Rembrandt works and has a unitary classification probability. The cloak, including the portion between the hand and the collar, was too visually simple to qualify for analysis. The tiles encompassing the strongly blue hand may simply have exerted undue influence on the lower portion of the collar due to the unclassified intervening area.

What about the book? Once the workshop door is figuratively thrown open and the authorship of assistants conceded, almost any region of the painting can plausibly qualify for a blue classification — particularly simple or mundane elements that artists have traditionally delegated to assistants. Though difficult to see, the entire portion of the hat that qualified for analysis is classified as dark blue, and this classification may have bled spuriously into the book as a result of tile averaging. Somewhat awkwardly placed and visually simple, the hat does seem like an element that Rembrandt may have left to another hand, or perhaps it was added later. But even if we assume the blue classification to be correct — by no means a certainty, since no system is perfect — other explanations are possible. A blue-classified element may have been painted by Rembrandt but in a somewhat unusual style that falls outside the neural network's training. It may have been heavily restored later. The contemporary digital photograph of the painting may have been retouched, e.g., to eliminate a patch of glare. These different possible sources of authorship coexist, with varying degrees of likelihood, until resolved by records or authoritative scholarship — if ever. Until then, their coexistence expands our understanding of the painting's possible histories and, to different degrees, challenges the traditional pull toward assigning a singular authorship to great works of art.

Our probability maps contrast with recent efforts to highlight the portions of an image that proved most important to a CNN classification. The "Grad-CAM" technique (Selvaraju et al., 2017), for example, produces "heat maps" that color-code how much attention the CNN paid to different regions. A Grad-CAM heat map of a tile containing the face from the *Johannes Wtenbogaert* portrait appears in Fig. 2.

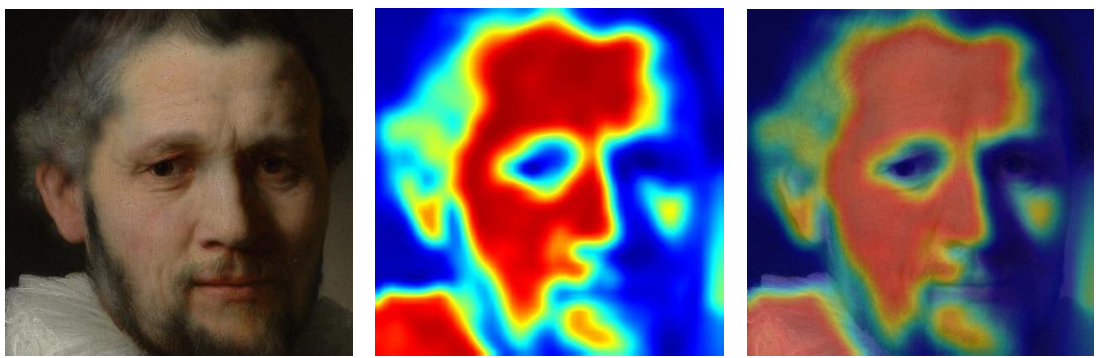

Fig. 2: (a) *Fragment of Portrait of Johannes Wtenbogaert*; (b) *Grad-CAM heat map of the fragment shown in* (a); (c) *Grad-CAM heat map superimposed on the fragment*

The Grad-CAM heat map reveals, unexcitingly, that the brightest areas of the painting attract the most CNN attention. These areas also exhibit greater visual diversity and, hence, higher image entropy, so they naturally play a larger role in classification. It is unlikely that lighting itself represents the basis for classification; many portraits classified differently by our system exhibit similar illumination patterns. Consequently, the Grad-CAM technique can tell us where a CNN looks, but cannot tell us what the CNN sees.

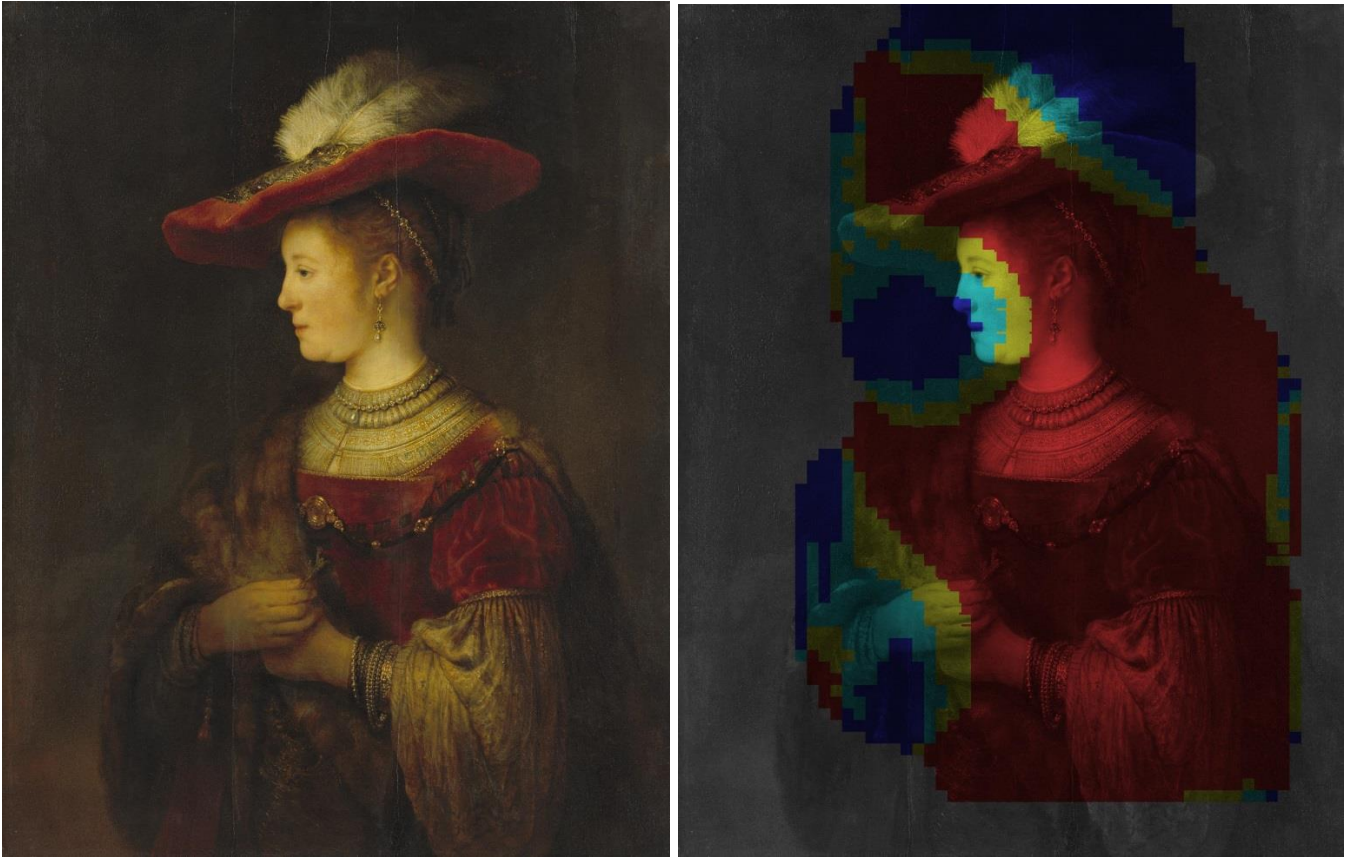

Fig. 3: *Portrait of Saskia (1642), Probability map*

Fig. 3 shows Rembrandt's portrait of his wife Saskia van Uylenburgh, believed to have been painted in 1642 — the year of Saskia's untimely death. Owned by the Altemeister Museum in Kassel, Germany, the painting has undergone substantial alteration since its creation. As detailed in van de Wetering (2017), the panel was reduced in size, "which also entailed overpainting the background as a whole. … Moreover, the face in the Kassel painting is virtually entirely overpainted; as with the background, a livelier *peinture* is hidden under this porcelain-like overpainting." The probability map generated by our CNN clearly highlights these alterations. Portions of the background visually busy enough to support classification are frequently mapped blue, as is a portion of the face, which does strike us as disappointingly featureless for Rembrandt. Blue-mapped portions of Saskia's hat may, once again, represent spurious spillover from the dark blue classification of the adjacent overpainted background. Our system's analysis suggests that the background to the right of Saskia's head received relatively little, if any, overpainting. Our model accords an overall probability of 0.78 to the painting, i.e., primarily Rembrandt but with subsequent revision.

The Kassel *Saskia* is often considered alongside his *Flora* (1660; Metropolitan Museum of Art, New York) (Fig. 4), which scholars believe Rembrandt based on the former.

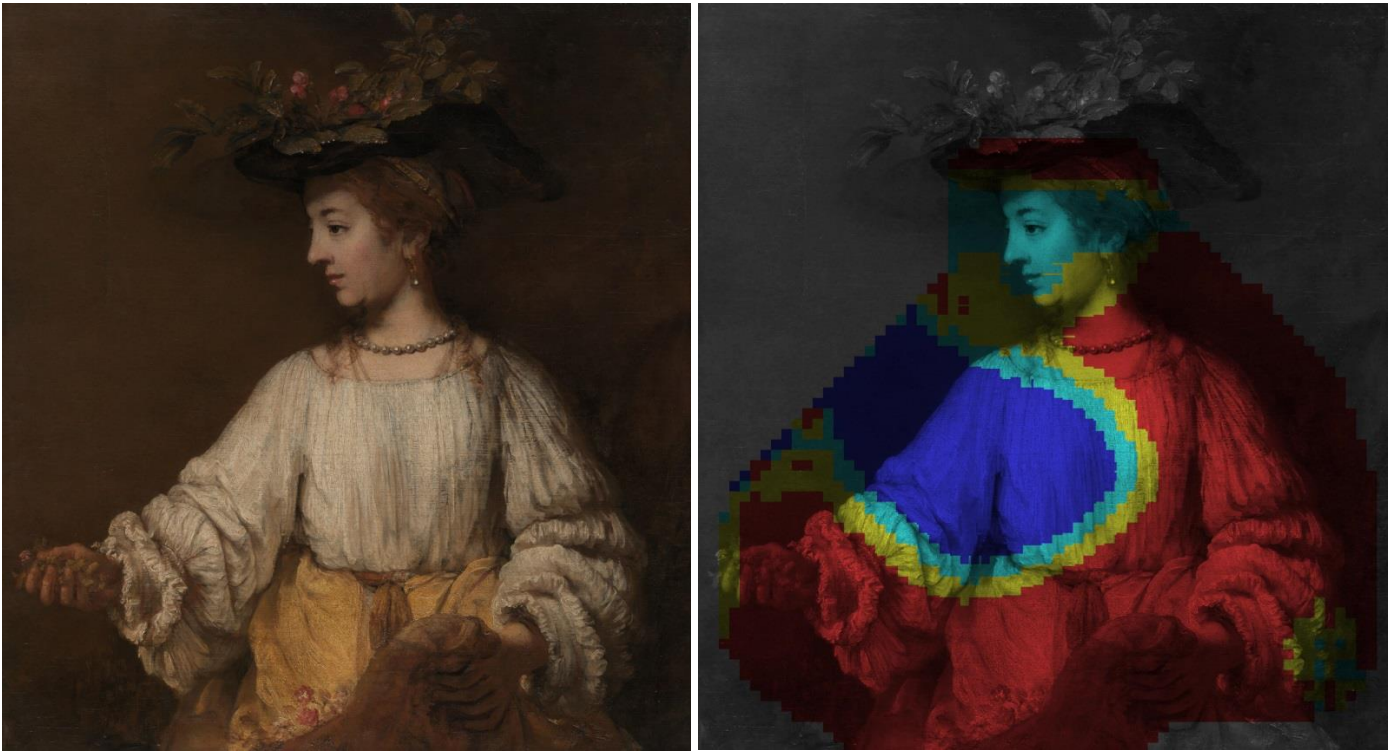

Fig. 4: *Flora (1660), Probability map*

Van de Wetering (2017) attributes this painting to "Rembrandt and workshop?" and perceives a "conspicuous difference in quality between different parts of the [painting]" — questioning the authenticity of the hand with the flowers.[3] Liedtke (2007) commented on the condition of the painting, noting that the "face, most of the blouse, and the skirt have lost much of their original surface in past restorations." Horizontal canvas threads peek through worn areas of paint across much of the figure. Liedtke also notes the existence of "pinpoint restorations on the neck" and removal in the 1970s of "an early repainting of the background." It is perhaps these restorations that cause our CNN to classify the face, which seems as featureless as the

---

[3] As can be seen from Fig. 4, our CNN classifies the hand and flowers as Rembrandt. To explore whether this result might have arisen spuriously as an accident of tile size, we isolated the hand and tested it, obtaining a classification that was decisely Rembrandt. Hands similarly isolated from the work of other artists were properly classified as not painted by Rembrandt.

Kassel *Saskia*, and a portion of the garment and background as not painted by Rembrandt; and to report only a tentative (gold-colored) Rembrandt classification along the neck.

As noted earlier, in five of six cases our system's classifications of paintings whose attributions have varied over time accord with the current scholarly consensus. The conspicuous exception is *Man with the Golden Helmet*, which was officially de-attributed by its owner, the Staatliche Museum in Berlin, in 1985. The celebrated depiction of a somber, elderly soldier in a spectacularly lit plumed helmet had consistently drawn large and appreciative crowds. "No other picture has played such an important role in the collection due to its enormous popularity and radiance," reads the museum's description of the painting, and following the de-attribution, the *New York Times* reported:

> For many visitors it ranked equal with the bust of Queen Nefertiti as the single most memorable object in what had once been the Kaiser-Friedrich Museum in central Berlin and was moved after World War II to the leafy and pacific suburb of Dahlem. All over the world there were houses and apartments in which it hung in reproduction, and at the postcard stand it was (as someone once said of commercial television in Britain) "a license to print money" (Russell, 1985).

The collective mourning that greeted the museum's decision continues to stir passions among the painting's many admirers. (We are among them.)

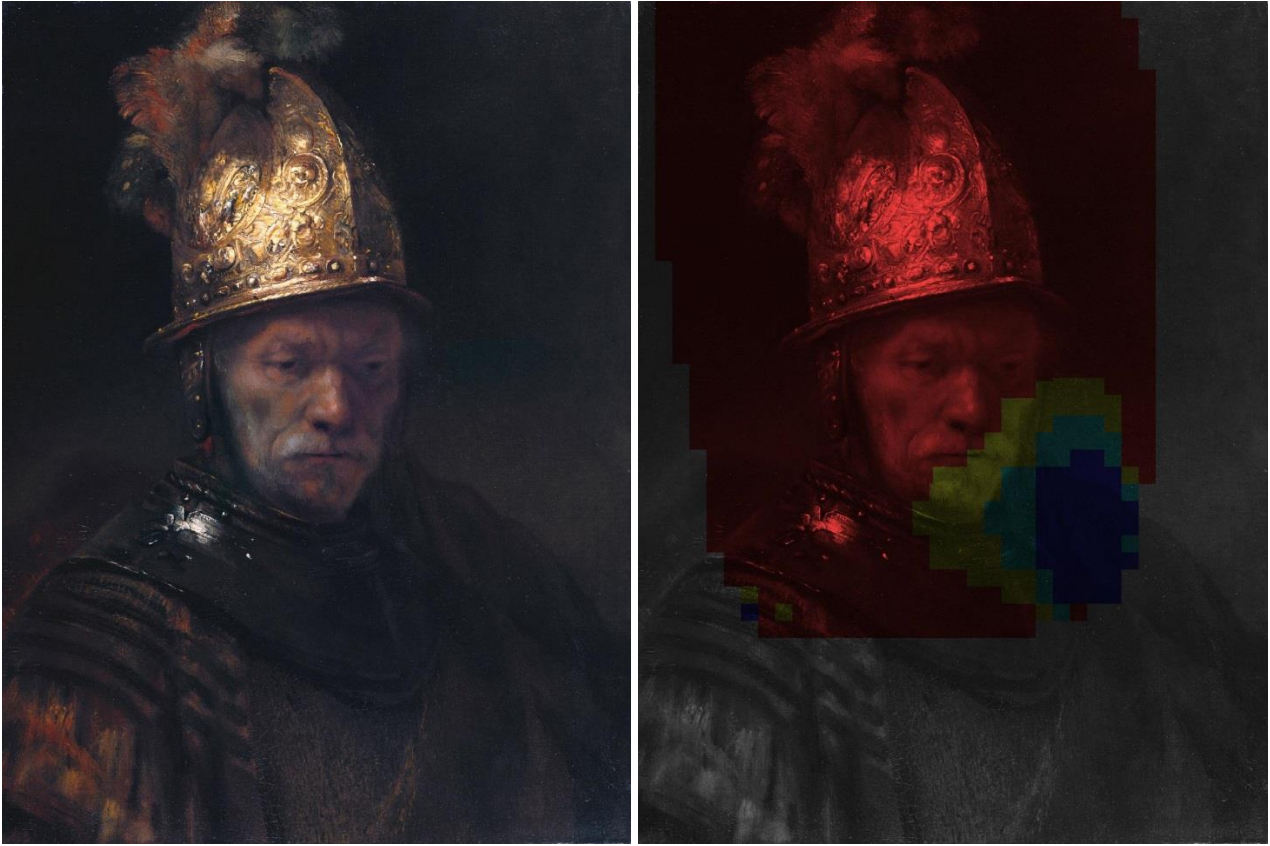

Fig. 5: *Man with the Golden Helmet (1650), Probability map*

We briefly discuss this painting in Frank et al. (2020), noting the attribution analysis by Staatliche Museum scholars: "In particular the thick application of paint to the helmet in contrast to the conspicuously flat rendering of the face, robe and background, which are placed adjacent to each other without a transition, does not correspond to Rembrandt's way of working" (Kleinert et al.). Given our CNN's poor classification accuracy for Rembrandts when analysis is confined to detail-level tiles, we modestly questioned the reliability of attribution judgments based on fine surface features. Here we further observe that our CNN, too, classifies the transition between the face and the robe as not Rembrandt (or, more accurately, not representative of Rembrandt according to the CNN's training). Nonetheless, the face and helmet are decisively classified as Rembrandt, particularly given the likelihood that the lower probabilities assigned to gold-colored portions of the face arise from spillover from the blue-colored robe; if tiles corresponding to the robe were removed from the analysis, the entire face would almost certainly be classified fully red.

The judgment of the Staatliche Museum may reflect the traditional tendency to assume a sole primary author, "the singularity of the master's touch" (Guichard, 2010). If elements of the painting depart from Rembrandt's known working style, in other words, it must not have been painted by Rembrandt. Our analysis suggests the possibility of multiple authors, with a narrower conclusion drawn from discrete stylistic departures: if we concede the propriety of classifying large, salient portions the painting as Rembrandt, those classifications are not necessarily undermined by adjacent non-Rembrandt classifications. Instead, the latter may suggest collaboration or delegation, later emendation, or restoration — or perhaps even an unusual experiment by the master himself, sufficiently distinct from his typical working technique to fool our system as well as the expert's eye. The robe is clearly rougher and less worked than the face and helmet, and while such treatment is frequently seen Rembrandt's later work, it would be unusual as early as the ascribed completion date of 1650.

## *Neural Network Analysis of Van Gogh Landscapes*

Although van Gogh worked alone, virtually eliminating ambiguities arising from mixed authorship, John Rewald, the distinguished historian of Impressionism and Post-Impressionism, remarked that van Gogh may well have been forged "more frequently than any other modern master" (Bailey 1997). Perhaps the most scandalous instance of fraud occurred in 1928, when the Berlin art dealer Otto Wacker exhibited a series of previously unknown van Gogh works. Wacker claimed to represent a mysterious Russian aristocrat on the run from the Bolsheviks, forced to sell his precious collection to save his family. Although the fakes fooled Baart de la Faille, author of the first catalogue raisonné of van Gogh's works, others questioned the works' authenticity as soon as the exhibit opened. Wacker was prosecuted and served 19 months in prison.

Debate over the authenticity of works attributed to van Gogh did not end with the Wacker affair. Scholars cite numerous reasons why the artist has been such tempting fodder for forgers. Foremost is, of course, money. Although van Gogh sold few paintings during his lifetime, his work was in high demand within just decades of his death, and prices for canvases and drawings by the artist climbed steadily for the next century. Today, prices fetched for his work regularly break auction records. Because van Gogh sent most of his paintings to his brother Theo during his lifetime, there are no dealers' records from that period.

Van Gogh wrote about specific works in his frequent correspondence with Theo and others, and his letters provide forgers with descriptions that can be used to create fakes — especially when authenticated works matching the artist's written descriptions are not otherwise known. Van Gogh's oft-noted practices of giving away paintings to friends and exchanging work with other artists can provide cover for a murky early provenance. And van Gogh did not always sign his work, so the absence of a signature might not raise the suspicions of a dealer or collector.

Academic scholars, professional connoisseurs, and Internet provocateurs have questioned various works, some quite famous, on stylistic or provenance grounds. Van Gogh's style and working methods were far from static, however, complicating judgments based on perceived technique or composition. His prodigious output and chaotic habits preclude any definitive catalogue raisonné.

The dataset we employed to train our CNN for van Gogh consisted of high-resolution images of 58 of his landscape paintings, selected to be representative of his evolving landscape painting styles, and 58 landscape paintings by other artists, intended to span a range of visual similarity to the van Gogh landscapes — from very close to evocative but readily distinguishable. Our objective was to train the CNN to make fine distinctions with good generalization properties and without overfitting to the training set.

The other artists whose work we selected for the dataset include older artists whom van Gogh admired and learned from (Charles-François Daubigny, Claude Monet, and Camille Pissarro); contemporaries of van Gogh (Emile Bernard, Paul Cézanne, Paul Gauguin, Georges Seurat, and Paul Signac); and later artists — especially the Fauves and Expressionists (André Derain, Erich Heckel, Wassily Kandinsky, Ernst Ludwig Kirchner, Gustav Klimt, Henri Matisse, Gabriele Munter, Emil Nolde, Max Pechstein, Karl Schmidt-Rottluf, and Maurice de Vlaminck) — whose use of color and untraditional brushwork to express the artist's emotional response to a subject seems indebted to van Gogh. We also included several paintings by Cuno Amiet (1868-1961), a Swiss painter who explored and worked in numerous styles over the course of his long career. While a member of the Pont-Aven School in the early 1890s, Amiet became an admirer of van Gogh's work, and many of Amiet's paintings from the next several decades seem to consciously incorporate many elements of the Dutch artist's style. In 1907, Amiet painted a widely known copy of van Gogh's 1890 canvas *Two Children* (private collection). The Van Gogh Museum acquired Amiet's copy in 2001.

Our test set consisted of 27 van Gogh landscape paintings and 24 paintings by the other artists listed above. We obtained maximum classification accuracy on the test set using 100×100-pixel tiles (88%) and 150×150-pixel tiles (90%). Accuracy decreased steadily at larger tile sizes. We interpret these results to suggest that van Gogh's uniqueness — what sets his work apart, for classification purposes, from imitators and the artists he influenced — is most pronounced at the brushstroke level. While other artists might indistinguishably imitate his compositions and large-scale features, in other words, they do not (or cannot) duplicate his brushwork. We found that classification error can be reduced further by combining the probabilities associated with the best-performing tile sets, since the different CNN models "see" differently scaled features; the combination appears to mitigate errors associated with each tile set considered separately. Rather than taking a straight average, we computed weights for the 100×100 and 150×150 probabilities that minimize the overall classification error, resulting in a final accuracy of 94% on our test set.

Why should allowing the CNN access to the compositional elements of a van Gogh landscape destroy predictive accuracy if larger tiles merely make more information available — i.e., allow the CNN to process larger visual motifs in addition to, not in lieu of, the smaller brushstroke details? Perhaps other artists structure landscape compositions in a manner similar enough to van Gogh to preclude visual differentiation at that scale, or perhaps van Gogh's own compositions are sufficiently eclectic as to have the same effect. We cannot tell because, like all CNNs, our system resists reverse engineering. For simple classification tasks such as determining whether an image contains a cat or a clock, the intermediate layers of processing can be analyzed to identify features (e.g., a cat's eyes and ears) that contribute most strongly to a classification prediction. What distinguishes an artist from his or her imitators, by contrast, is infinitely more subtle, and the predictively significant features do not cohere into any sort of unifying theme.

The probability maps derived from van Gogh landscapes differ from the Rembrandt maps in two important ways. First, in many instances even less of the image is analyzed despite the high classification accuracy. Second, correctly classified van Gogh paintings tend to contain fewer oppositely classified regions; their overall probability scores typically exceed 90%. This again conforms to expectation, since van Gogh worked alone, so blue-classified regions of his paintings almost certainly represent errors. *Farmhouse in a Wheatfield* (1888, Van Gogh Museum, Amsterdam) exemplifies the typical pattern.

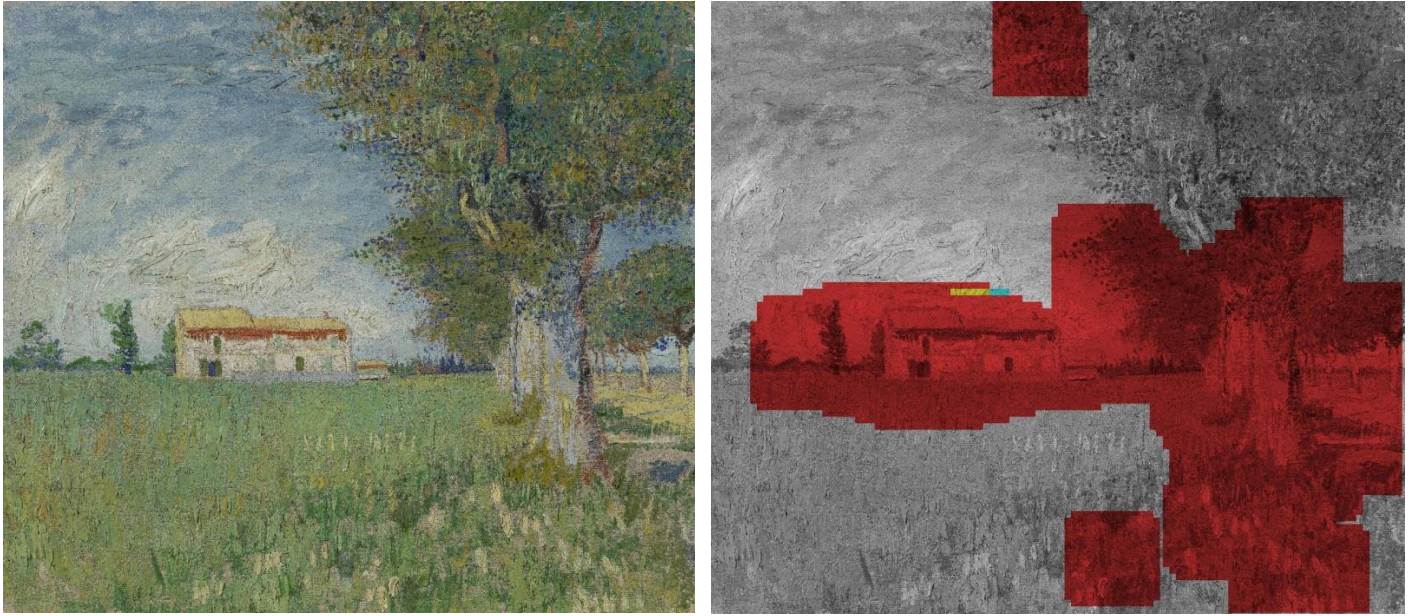

Fig. 6: *Farmhouse in a Wheatfield (1888), Probability map (150×150 tiles)*

A less-decisively classified painting was *Green Wheatfield with Cypress* (1889, National Gallery, Prague), shown with its probability map in Fig. 7.

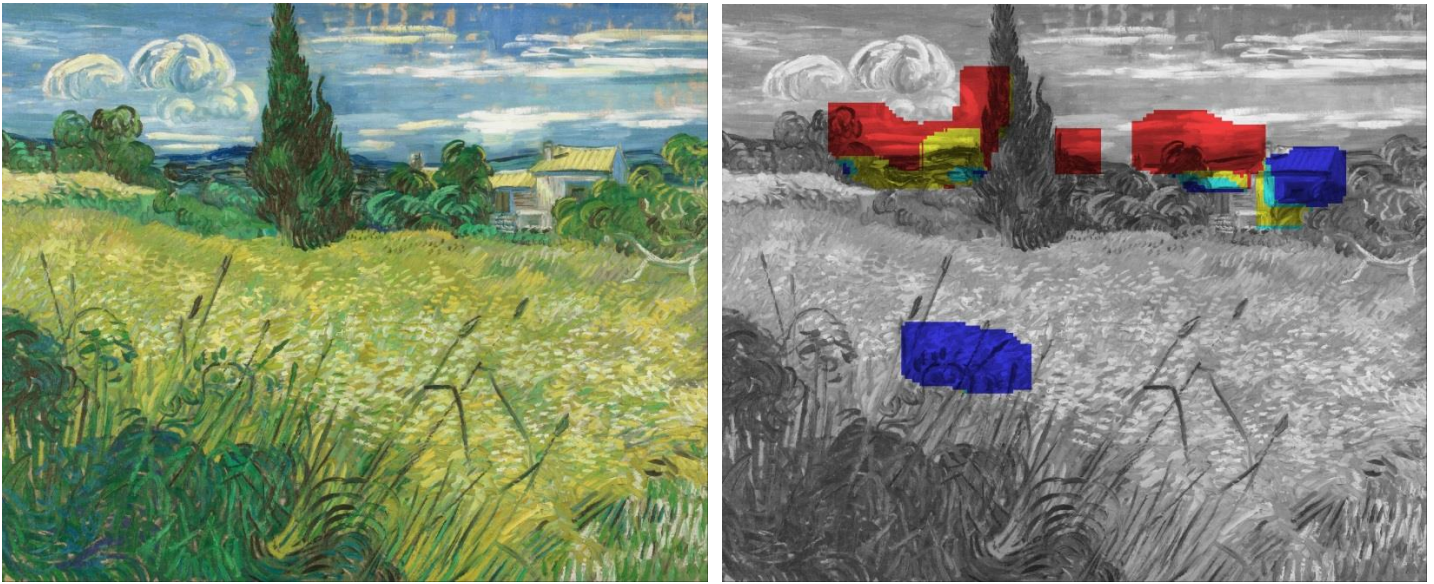

Fig. 7: *Green Wheatfield with Cypress* (1889*), Probability map (150×150 tiles)*

Our CNN examined a quite small proportion of this painting and prominently classified the house as not by van Gogh. While the house may have been *painted* by van Gogh, perhaps its brushwork is not *characteristic* of van Gogh — at least insofar as our training set is concerned. That is, more non-van Gogh landscapes in our training set had passages with this brushstroke style than did van Gogh paintings.

Once again, we tested van Gogh works that have been subject to attribution controversy to see whether our CNN's classification would match the current scholarly consensus. In addition, we tested a work painted to resemble van Gogh by the famous forger John Myatt. All four of these works and their associated probabilities with 150×150-pixel tiles are shown in Fig. 8.

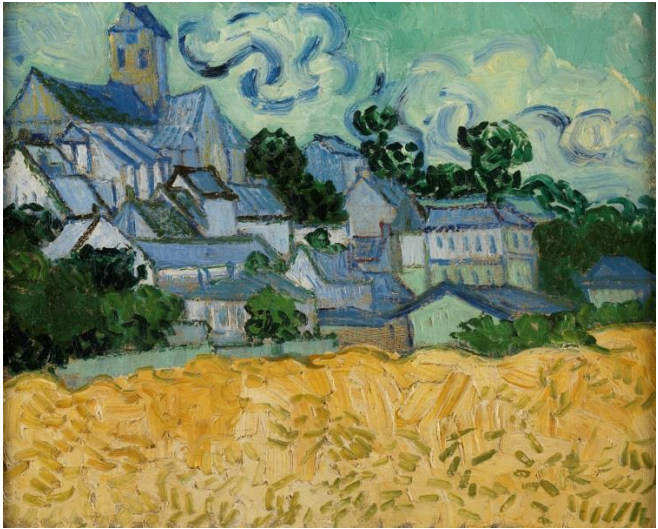
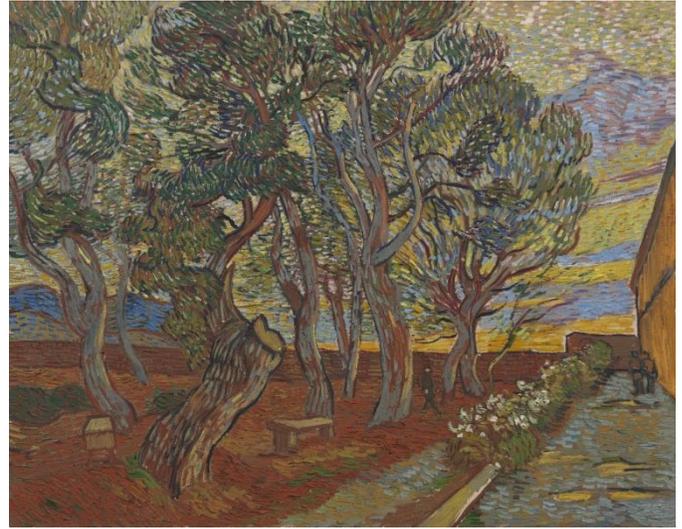
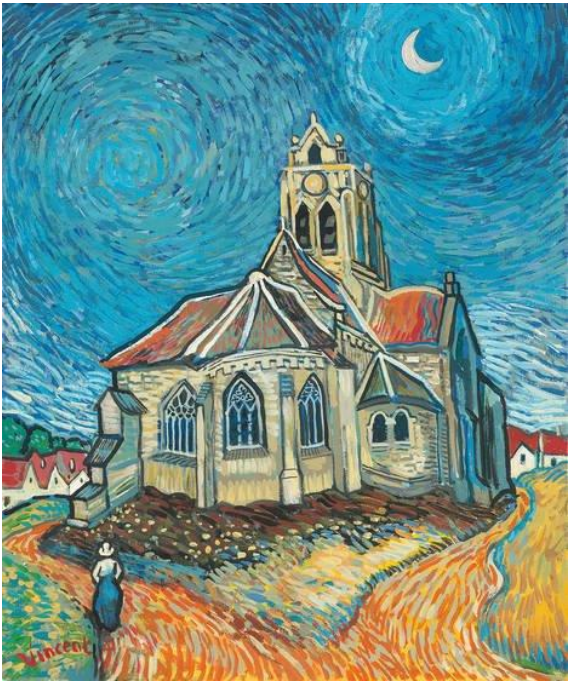
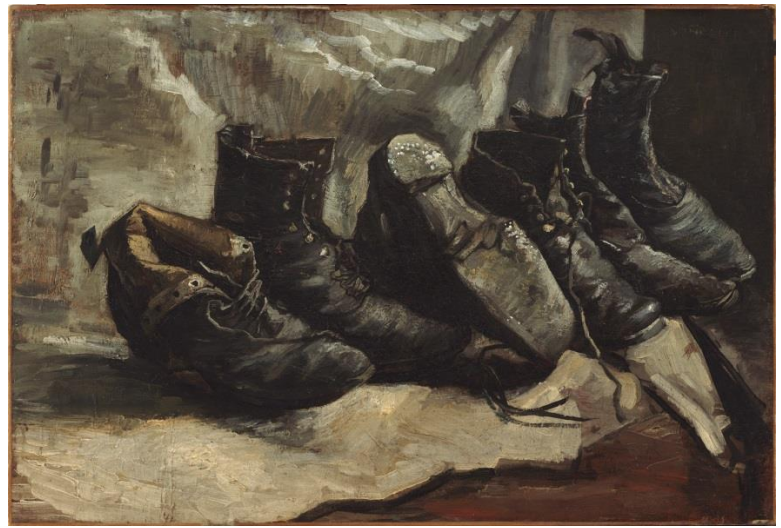

Fig. 8: Clockwise: *View of Auvers-sur-Oise* (1890; Museum of the Rhode Island School of Design, Providence, RI), 0.90; *Garden of the Asylum* (1889; Van Gogh Museum, Amsterdam), 0.98; *Three Pairs of Shoes* (1886-87, Fogg Art Museum, Cambridge, MA), 0.54; John Myatt, Church at Auvers (2017), 0.29

Each painting was classified correctly. Notably, *Three Pairs of Shoes* is classified as having been painted by van Gogh with considerably lower probability than the other two van Gogh works. In assigning meaning to this result, it is important to remember that we trained our CNN on van Gogh *landscapes*, and much of *Three Pairs of Shoes* differs considerably in style and content from our training set. While it is perhaps unsurprising that van Gogh's detail-level "signature" persists across genres and enough similarities exist at the brushstroke level to support a successful overall classification, cross-genre analyses like this have intrinsic risks, which we plan to explore in future work.

*Conclusion*

Accepted scientific techniques of artwork authentication such as X-ray analysis, dendochronology, scanning-electron microscopy, and radiocarbon dating tend to answer specific questions — when the artwork (or its constituents) was made and whether hidden layers of prior work underlie the visible surface. Only in the latter case can the results tell a creation story, and then only in the unusual case where multiple hidden layers can be detected, resolved, and understood as an accretive sequence. Our CNN approach, by contrast, always tells a story — or, rather, invites a story to be told. Often a probability map will contain tantalizing clues that plausibly cohere into a creation narrative. It is important, in assigning meaning to a probability map, to resist allowing the coherence of the story to inflate the accuracy of the technique or the reliability of conclusions. From cosmologists who judge the validity of a hypothesis by the elegance of its mathematics to the "just so" stories sometimes proposed to explain evolutionary adaptations, the temptation of the narrative explanation is strong. We understand the world through stories. To constitute a valid explanation, however, the story must have predictive utility — easier to test in the local world of biology than the infinite universe of cosmology or, in our case, a solitary map of probabilities for one unique work.

Any hypothesis regarding a painting's authorship must therefore be grounded in traditional scholarship and scientific analysis, and qualified by inherent limitations in accuracy. At the same time, judgments should not be cabined by fealty to conventional notions of singular authorship. We believe that CNN analysis of two-dimensional artwork can help resolve basic questions of authenticity as well as more subtle nuances that trace the lived history of the work. Though computational, the technique is not mechanical, but is as firmly tied to its experience — i.e., its training — as is a human expert to his or her own. The objective in any "deep learning" exercise is to achieve generalization beyond the training images, but results can never be foolproof because the extent of that generalization is unknowable; the full range of an artist's style as well as the limitless ways it can be imitated will always surpass the limits of training. It may take more than one voice to tell the full creation story of a masterpiece, and we hope our work can contribute to the narrative.